\documentclass{NLE}
\usepackage{amsmath,bm}
\usepackage{graphicx}
\usepackage{natbib}
\ifpdf%
\usepackage{epstopdf}%
\else%
\fi

\usepackage{multirow}
\usepackage[export]{adjustbox}
\usepackage{setspace}
\usepackage{subfig}

\usepackage{graphicx}
\usepackage{amssymb}
\usepackage{breqn}
\usepackage{hhline}
\usepackage[ruled,vlined]{algorithm2e}

\usepackage{amsthm}
\usepackage{multirow}
\usepackage{floatrow}
\usepackage{float}
\usepackage{algorithmicx}
\SetKwInput{KwInput}{Input}  
\SetKwInput{KwOutput}{Output}
\DeclareMathOperator*{\argmax}{arg\,max}

\DeclareMathOperator*{\argmin}{arg\,min}
\newcommand{\norm}[1]{\left\lVert#1\right\rVert}
\newcommand{\matr}[1]{\mathbf{#1}}

\DeclareMathOperator*{\transpose}{T}

\DeclareCaptionLabelFormat{nospace}{#1 #2}
\begin{document}
\label{firstpage}

\lefttitle{CPM-based Modelling in NLU}
\righttitle{Natural Language Engineering}

\papertitle{Article}

\jnlPage{1}{00}
\jnlDoiYr{2019}
\doival{10.1017/xxxxx}

\title{Convex Polytope Modelling for Unsupervised Derivation of Semantic Structure for Data-efficient Natural Language Understanding}
\begin{authgrp}
\author{Jingyan Zhou$^1$, Xiaohan Feng$^1$, King Keung Wu$^2$, Helen Meng$^1$}
\affiliation{$^1$Dept. of Systems Engineering \& Engineering Management, The Chinese University of Hong Kong\\
  $^2$ SpeechX Limited\\
  \email{\{jyzhou, xhfeng, hmmeng\}@se.cuhk.edu.hk, kkwu@speechx.cn}}
\end{authgrp}
\history{(Received xx xxx xxx; revised xx xxx xxx; accepted xx xxx xxx)}

\begin{abstract}
    Popular approaches for Natural Language Understanding (NLU) usually rely on a huge amount of annotated data or handcrafted rules, which is laborious and not adaptive to domain extension. We recently proposed a Convex-Polytopic-Model-based framework that shows great potential in automatically extracting \textit{semantic patterns} by exploiting the raw dialog corpus. The extracted semantic patterns can be used to generate semantic frames, which is essential in assisting NLU tasks. This paper further studies the CPM model in depth and visualizes its high interpretability and transparency at various levels. We show that this framework can exploit 
   semantic-frame-related features in the corpus, reveal the underlying semantic structure of the utterances, and boost the performance of the state-of-the-art NLU model with minimal supervision. We conduct our experiments on the ATIS (Air Travel Information System) corpus.
\end{abstract}
\maketitle

\section{Introduction}

Task-oriented dialog system is receiving increasing attention in both research and industrial community.  It can serve as a natural interface to help users achieve certain goals through natural conversation and has been adopted by many applications such as personal assistants, e-commerce sales assistants, etc.

\begin{figure}[thb]
    \centering
    \includegraphics[scale=.35]{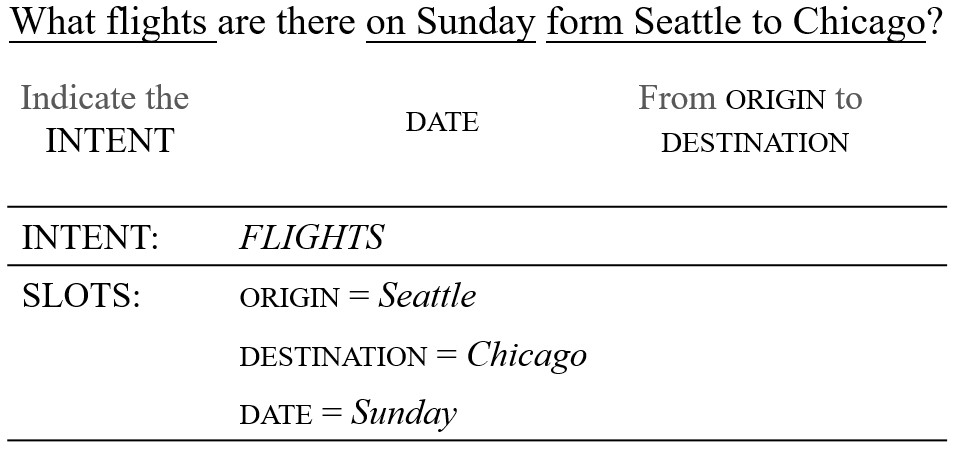}
    \caption{An utterance from the ATIS dataset inquiring about flight information, together with the corresponding semantic frame.}
    \label{fig:frame-example}
\end{figure}

To assist users to accomplish certain tasks, the system needs to understand the user’s utterances and extract necessary information from them, the process of which is usually referred to as Natural Language Understanding (NLU).
The extracted information is commonly represented as a structured form called the \textsc{semantic frame}, which involves user's intent and corresponding slot-value pairs. Fig. \ref{fig:frame-example} shows an example of a labelled user utterance from the Air Travel information Systems (ATIS) dataset \citep{price1990evaluation}, \textit{`What flights are there on Sunday from Seattle to Chicago'}, and its corresponding semantic frame \textit{`FLIGHTS (ORIGIN=Seattle, DESTINATION=Chicago, DATE=Sunday)'}. In this example, the intent of this user query is \textit{FLIGHTS}, and the corresponding slots are \textsc{origin, destination, date} and \textsc{time}.
Note that for a certain domain, such as the air travel information domain of ATIS, the set of possible intents and slots are usually finite but may vary in size due to extension of domain. 

A common approach towards NLU is the rule-based method, where domain experts define the semantic frames, design a set of heuristic rules to locate intents and slots \citep{saini2019linguistic}.
However, the grammar rules have limitations in modelling the complex structures of real-world dialogs as they are based on the syntactic, semantic and discourse information of the utterances. Hence, a complex dialog system may require a sizeable number of rules to function, which is difficult to design manually. 
Another popular approach is machine learning-based NLU \citep{firdaus2020deep, liu2016attention, chen2016knowledge} which requires a sizeable number of utterances with well-labelled intents and slots-value pairs as training data for supervised learning.
Although the neural-based supervised learning model has achieved the state-of-the-art NLU performance, it acts like a black box that reveals limited clues during the inference process of filling the semantic frame. 
Note that both approaches mentioned above require experts to either design the rules or label the training data, which is labour-intensive and thus costly. Besides, they are not adaptive to domain extensions, especially when the domain changes actively.

This paper presents an approach that can derive the \textit{semantic patterns} and reveal the underlying \textit{semantic structures} of the utterances in dialog interaction data, which only requires a corpus of sentences or transcribed utterances with minimal labelling.  This is aimed to facilitate natural language understanding with little supervision. 

We observe that in task-oriented dialogs, the information related to various intents and slots tend to be embedded in certain word patterns (e.g. phrases) that appear frequently. 
For example, in the ATIS corpus, the \textsc{origin} slot usually appears in the user’s query in the format of ‘from \textit{city-name}’, e.g., ‘from San Francisco’.
Note that here we delexicalize the tokens with meaningful labels according to their categories, e.g., replacing ‘San Francisco’ with ‘\textit{city-name}’.
Also, these patterns may not necessarily be consecutive words.  For instance, the frequently observed pattern ‘show me ... flights’ indicates the user intent \textit{FLIGHTS}, where ‘...’  means that any words can appear within this pattern and do not affect its meaning.
Hence, we can leverage such \textit{semantic patterns} in deriving the semantic frames of the utterances. 

Furthermore, we also notice that an utterance is usually composed of several patterns, e.g., a user’s query for searching flights usually contains a combination of patterns related to the intent (i.e. \textit{FLIGHTS}) and slots (e.g. \textsc{origin, destination, date} and \textsc{time}). Therefore, revealing such combinations equals to constructing the \textit{semantic structure} of the utterance and can help to understand the utterance.

Based on the observations above, we devise a data-driven, unsupervised yet interpretable framework based on the Convex Polytopic Model (CPM) \citep{wu2018topic}, which can (i) automatically discover potential semantic patterns by analysing raw dialog data; (ii) facilitate analyses and visualization of the underlying semantics structures 
and (iii) allow injection of CPM-extracted features into a neural-based model such that the incorporated knowledge can help lower the number requirements of labelled training data. 

CPM is an unsupervised geometric algorithm that can extract the key semantic concepts from corpora of short texts \citep{wu2018topic}, which was previous applied to the discovery of information topics from a corpus of short documents. This paper applies a similar algorithm to the utterances from a dialog corpus by first projecting the utterances to a low-dimensional affine subspace via Principal Component Analysis (PCA), and then enclosing the projected points within a compact convex polytope. The CPM algorithm can effectively discover the semantic patterns (exhibited as polytope vertices) that are closely related to the semantic frames and can decompose all utterances in the dataset into combinations of the semantic patterns.

We apply CPM on the ATIS dataset to demonstrate its ability in extracting useful features (i.e. semantic patterns and structures) for achieving NLU. 
First, we show that the polytope vertices are closely related to the semantic patterns by comparing the results with human-labelled data.
Second, we analyse the semantic structure of individual utterance from two perspectives: 
(i) \textit{compositional analyses} -- as each utterance point in the polytope can be represented as a convex combination of vertices, the combination coefficients can thus reveal the contribution of the semantic patterns to the utterance; 
(ii) \textit{word-word correlation analyses} -- the word-word correlation within the utterance can be discovered based on the word vectors using semantic patterns as the features, which reveals the internal semantic structure. 
Third, we further incorporate the CPM-discovered features to guide the neural-based model in the example task of slot filling, where we show that such incorporation can achieve significantly better performance when there is limited amount of labelled data.

Our contributions are three-fold: 
(1) we propose a novel, unsupervised method via CPM to extract semantic patterns from task-oriented dialog corpus and discover the underlying semantic structures of the utterances;
(2) we demonstrate the high interpretability and transparency of the CPM model via various visualization techniques; 
(3) we verify that the CPM extracted features can boost the performance of the NLU models in cases with limited labelled data, while enhancing the interpretability of the neural-based models for slot filling.

\section{Literature Review}
\textbf{Dialog data mining.} Recently, there is a series of active research in leveraging raw dialog datasets from human-human or human-machine conversations to facilitate NLU.  \cite{haponchyk2018supervised, perkins2019dialog, chatterjee2020intent} apply clustering algorithms to implement an expert-in-the-loop approach to assist labeling of semantic frames. These approaches require some seed data, and the results are efficient but noisy. \cite{mallinar2019bootstrapping} aims to provide a more human-friendly interface for expert annotators and adopts data programming techniques to expand the labels to more data. These approaches try to produce more semantic clusters to capture the users’ intents in their queries.
In addition to the user’s intents, key semantic concepts, often referred to as `slots', are also important for understanding the users’ queries.   Our previous work 
\citep{zhou2021automatic} has shown that the CPM-based geometrical framework can be used to extract semantic structures from raw conversational data automatically.
The current paper focuses on extracting the semantic features and approaching NLU or assisting current NLU models with the extracted features.

\noindent\textbf{Natural language understanding.} 
Conventional approaches in rule-based parsing for semantic information is still being used, such as the E-commerce dialog system, AliMe Assist \citep{li2017alime}, which adopts a business rule parser for NLU.  However, rule-based approaches are giving way to statistical approaches, especially for some deployed personal assistants like Google Assistant \citep{google2018}, Microsoft Cortana \citep{gao2019neural}, etc.
Furthermore, the recent trend of using machine learning approaches to NLU often organize the problem into two sub-tasks: (i) an utterance classification task for intent identification, and (ii) a sequence labeling task for slot-filling.  Classifiers such as support vector machines (SVM) and deep neural networks \citep{sheil2018predicting} have been applied to the intent classification task. For the slot-filling task, previous research has applied sequence tagging techniques such as Conditional Random Fields \citep{tang2020end} and neural approaches such as Recurrent Neural Networks (RNN) \citep{mesnil2014using}. More recent works apply joint modelling of the two tasks, such as \citep{liu2016attention} and \citep{wang2018bi}.

In this work, we conduct a CPM-guided NLU experiment on the RNN-based model, which is widely adopted and studied in this field. Here we employ the model proposed by \cite{liu2016attention} as an example to demonstrate the synergy between CPM and neural models on NLU tasks.

\noindent\textbf{Knowledge-guided NLU.}
There is an emerging trend of injecting knowledge to guide machine learning-based NLU models to enhance their interpretability, as well as reduce their requirements for well-labelled data.  \cite{chen2016knowledge} introduces the dependency parser result to guide the RNN for slot filling, which achieves good performance with only a small amount of well-labelled data. 
The popular pre-trained models like BERT, which are proven to have learned rich syntactic knowledge from a huge amount of training data \citep{manning2020emergent}, have also been used for NLU tasks and achieved significant improvement \citep{chen2019bert}.

Since the approaches reviewed above generally capture generic syntactic knowledge learned from the datasets that are unrelated to the application domain, they lack task-specific semantic knowledge. Therefore, this work applies CPM-extracted knowledge that aims to closely associate with the semantic frames, as well as maintain human interpretability in the development of NLU models.

\section{Convex Polytopic Model}
\subsection{Formulation of CPM}

The approach of CPM for NLU is formulated as three key steps: (i) Represent each utterance in the dialog corpus by a sum-normalized term frequency vector; (ii) Project all the utterance vectors onto a low-dimensional affine subspace via Principal Component Analysis (PCA); and (iii) Generate a Minimum Volume Simplex (MVS), a special type of convex polytope, to enclose the projected utterances. The details of each step will be explained in the following subsections.
\subsubsection*{Step 1: Represent each utterance as a sum-normalized term frequency vector}

Given a task-oriented dialog corpus of $M$ utterances, we pre-process it by removing the words that occur with frequency lower than a threshold (set at 2 in our experiments), converting all capital letters to the lowercase alternative, and building a vocabulary of size $N$. 
We then represent each utterance by an $N$-dimensional vector with each entry as the occurrence count of the corresponding word divided by the total number of words in the utterance, which is referred to as a sum-normalized term frequency vector, where the sum equals one. 
The term frequency vector has traditionally been applied as document representation in natural language processing and information retrieval. 
Such representation discards the grammar and word order while largely preserving the semantics of the document. We therefore expect that in the use case of dialog corpus, it can also preserve the semantics of the utterance.

Geometrically, the utterance vector can be considered as a point on a $(N-1)$-dimensional sum-to-one hyperplane in the non-negative orthant of the $N$-dimensional space. Note that the sum-to-one hyperplane in the non-negative orthant forms an $(N-1)$-dimensional regular simplex with vertices on the axes and the mean position of all the vertices as the vertex centroid, denoted as $\matr{c_s}$ here. 
As the vector entry is computed by dividing the total number of words, the vector of a shorter utterance tends to be sparser and its non-zero entries have larger values, hence its associated point tends to lie in the peripheral region and be farther from $c_s$. Besides, the Euclidean distance between two utterance points characterizes the difference in semantics between their corresponding utterances, as the utterances with similar semantics usually appear to have more overlapping words and thus the difference between their vectors is also smaller.

\subsubsection*{Step 2: Project utterance vectors onto a low-dimensional affine subspace via PCA}

We aim to capture the \textit{important semantic information} for building the semantic frames by filtering out less important features. 
Hence, we propose to remove the components with insignificant effect using PCA, by discarding the small principal components. It is geometrically equivalent to embedding the utterances into a low-dimensional affine subspace spanned by the principal axes computed by PCA. The process of PCA reduces the dimensionality by maximizing the variance which preserves large pairwise distances and thus the important semantic features (i.e. distinguishing features) of the utterances can remain in the projected points. We refer to these points as \textit{projected utterance points} to distinguish them from the original utterance points represented by the term frequency vector in Step 1.

Mathematically, the PCA process can be formulated as follows. We stack all the term utterance (column) vectors in the corpus to form a term-utterance matrix $\matr{X}\in\mathbb{R}^{N\times M}$. Let the column vector $\bar{\matr{x}}\in\mathbb{R}^{N}$ be the column average of $\matr{X}$ and repeat $\bar{\matr{x}}$ $M$ times to form $\bar{\matr{X}}\in\mathbb{R}^{N\times M}$. Then, we apply Singular Value Decomposition (SVD) to find the closest $R$-dimensional subspace that approximate $(\matr{X}-\bar{\matr{X}})$, where $R$ is the dimension of the low-dimensional subspace. By keeping the first $R$ eigenvectors of $(\matr{X}-\bar{\matr{X}})(\matr{X}-\bar{\matr{X}})^\transpose$, we obtain an orthonormal basis $\matr{U}\in\mathbb{R}^{N\times R}$. The utterance vectors can then be projected onto the $R$-dimensional utterance subspace spanned by the basis $\matr{U}$. The projected utterance points, denoted by $\{\matr{p}_i\}^M_{i=1}$, can be represented in the original $N$-dimensional subspace and be computed by the following formula:
\begin{equation}
	\label{eq:P}
    \matr{P} = \bar{\matr{X}} + \matr{U}\matr{U}^\transpose (\matr{X}-\bar{\matr{X}}).
\end{equation}
where the $\{\matr{p}_i\}^M_{i=1}$ are the columns of $\matr{P}\in\mathbb{R}^{N\times M}$. Note that the columns of $\matr{P}$ still have the \textit{sum-to-one} property, and consequently, the \textit{projected utterance points} are still on the \textit{sum-to-one hyperplane}. 
As the PCA transformation is essentially an affine transformation on the coordinates, the pointed utterance points that are away from the centroid also lie in the peripheral region before projection, but this is not guaranteed vice versa.

We highlight here the difference between the proposed PCA projection from the traditional algorithm of Latent Semantic Analysis (LSA) applied in information retrieval. In PCA, we aim to find the axes of an ellipsoid as the principal components by fitting the original utterance points in $\matr{X}$ into an $R$-dimensional ellipsoid. Note that the subspace spanned by the principal component axes is an affine subspace as we have subtracted $\matr{X}$ by its average $\bar{\matr{X}}$. However in LSA, the matrix $\matr{X}$ is neither sum-normalized nor subtracted by the average, and hence latent semantic space obtained by LSA is not an affine subspace. The key difference between the \textit{affine utterance subspace} and the \textit{latent semantic space} is that the former measures the difference of utterances by the spatial distance of their utterance points, while the latter measures with their included angle. We will show how this feature of the affine utterance subspace is useful for finding meaningful features for the dialog corpus in the next step.
 
From the perspective of a task-oriented dialog, the semantics of the utterance are captured by the \textit{semantic frame} which consists of the intent, together with slot-value pairs. As the process of PCA preserves the important semantics features in utterance by its associated location in the space spanned by the principal components, we expect that the distance between the resulting utterance points can distinguish the difference in the semantic frames of the utterances.

\subsubsection*{Step 3: Generate a Minimum Volume Simplex (MVS) to enclose the projected utterance points}

We aim to extract the semantic features that can be \textit{interpretable} from the projected utterance points based on geometric properties. 
Although PCA can preserve the most important semantics of a given corpus, the principal component axes may not be directly associated with linguistic concepts and thus uninterpretable. Also, the orthogonal property of the principal components may not always be compatible with the nature of semantic features, as the features are often not completely independent.
As we discussed previously, semantic patterns are critical components in forming semantic frames. Hence, we propose to apply the MVS algorithm to extract such semantic patterns as \textit{interpretable} features.

To represent the combination of semantic patterns in the mathematical way, we attempt to represent each \textit{projected utterance point} (i.e. a point in the subspace that represents an utterance in the corpus) by a \textit{unique} convex combination of the \textit{semantic patterns} (which will be defined in the following). 
Geometrically, it is equivalent to generating a simplex to enclose all the projected utterance points, as any point within the simplex can be uniquely represented by a convex combination of vertices of the simplex. 
These vertices can thus be considered as representations of \textit{semantic patterns}. 
However, there can be infinite many possible choices of an enclosing simplex for a point set.
Hence, we need to set a certain constraint to ensure the semantic patterns captured by the vertices are similar to the actual underlying patterns in the dataset. 
We propose to apply the minimum volume constraint to best fit the finite region occupied by the finite set of projected utterance points. 
This is to guarantee that the enclosing simplex fully covers the scope of the given corpus, with as little excessive area, corresponding to the exterior scope of the corpus, as possible. 
Note that a simplex in $R$-dimensional space has a fixed number of vertices of $R+1$ which means that the dimension of the affine subspace obtained by PCA controls the number of \textit{semantic patterns} that can be extracted.
The detailed procedures of finding the MVS will be discussed in Section \ref{ssec:MVS}.

The reason of choosing the convex combination approach instead of the conventional clustering methods to extract interpretable semantic patterns is two-fold. 
First, clustering methods which try to cluster points with similar features usually do not effectively take into account the combination property of semantic patterns that are exhibited in the utterances of task-oriented dialog corpus. 
Second, as mentioned earlier, the sum-normalization forces the points for shorter utterances to be farther from the centre, and shorter utterances tend to have fewer semantic patterns (or possibly just a single pattern). 
Hence, key semantic patterns should be lying around the peripheral region and even outside the point set. 
Therefore, it is reasonable to apply convex combination to obtain the simplex such that its vertices serve as extreme points that capture the key semantic patterns.

\subsection{Finding the Minimum Volume Simplex}
\label{ssec:MVS}

This section describes the detailed formulation of MVS approach by first introducing the convex polytopic representation and then formulating the optimization problem of finding the MVS.

\subsubsection{Convex polytopic representation}
\label{sec:cp}
The \textit{convex polytopic representation} of the \textit{projected utterance point} $\matr{p}_i$ is defined as the convex combinations of $\matr{v}_j$:
\begin{equation}
	\label{eq:eq1}
    \matr{p}_i=\sum_{j=1}^K a_{ij} \matr{v}_j,
\end{equation}
where $K$ is the number of vertices of the convex polytope and $a_{ij}\geq 0$ (non-negativity condition) and $\sum_{j=1}^K a_{ij}=1$ (sum-to-one condition), for $i=1, \ldots, M$, are the convexity conditions. For all the points $\{\matr{p}_i\}^M_{i=1}$ from the corpus enclosed by a convex polytope with vertices $\{\matr{v}_j\in\mathbb{R}^N\}^K_{j=1}$, (\ref{eq:eq1}) can thus be written in matrix form:
\begin{equation}
	\label{eq:eq2}
    \matr{P}=\matr{V}\matr{A},
\end{equation}
where $\{\matr{v}_j\}^K_{j=1}$ form the columns of $\matr{V}\in\mathbb{R}^{N\times K}$ and $\matr{A}\in\mathbb{R}^{K\times M}$ with entries $a_{ij}$. Note that all columns of $\matr{A}$, denoted by $\matr{a}_1, \ldots, \matr{a}_M$, are sum-to-one and non-negative.

\subsubsection{Finding MVS as an optimization problem}

To generate the MVS enclosing a point set lying on a $R$-dimensional subspace, we can formulate it as an optimization problem using the convex polytopic representation. Note that the number of vertices in the MVS is $K=R+1$. Consider the projected utterance points of matrix $\matr{P}$ is known, we apply the algorithm originally proposed in \citep{li2008minimum} to find the MVS. The problem of MVS optimization can be expressed as follows:
\begin{equation}
	\label{eq:eq3}
    \textbf{P}=\textbf{VA} \quad s.t. \quad \textbf{A} \succeq 0 \quad and \quad \textbf{1}_K^\transpose \textbf{A}=\textbf{1}_M^\transpose
\end{equation}
where $\textbf{1}_K$ denotes the column vector of dimension $K$ with all entries equal \textit{one}.

Note that $\textbf{V}\in\mathbb{R}^{N\times K}$ is not invertible. Also, $\textbf{P}\in\mathbb{R}^{N\times M}$  are PCA-projected utterance points which have $R$ degrees of freedom. Therefore, we first perform coordinate transformation and denote $\tilde{\textbf{P}}\in\mathbb{R}^{K\times M}$ and $\tilde{\textbf{V}}\in\mathbb{R}^{K\times K}$ as the transformed version of $\matr{P}$ and $\matr{V}$, respectively. 
Then we need to define the transformation matrix. The PCA-projected $R$-dimensional subspace is spanned by the basis $\textbf{U}\in\mathbb{R}^{N\times R}$. Also, the origin $o_r$ of the $R$-dimensional affine subspace is different from the origin $o_n$ of the original $N$-dimensional subspace. Therefore, we need to add one dimension for displacement from $o_r$ to $o_n$, which is  the orthonormal component of $\bar{\matr{x}}$: $\matr{u_{orth}}=\matr{u_0}/\|\matr{u_0}\|$ where $\matr{u_0}=\bar{\matr{x}}-\matr{U}\matr{U}^\transpose \bar{\matr{x}}$.
Hence, the transformation matrix $\tilde{\textbf{U}}$ can be defined as $\tilde{\textbf{U}} = [\matr{U}\; \matr{u_{orth}}]\in\mathbb{R}^{N\times K}$, $\tilde{\textbf{P}}=\tilde{\textbf{U}}^\transpose\matr{P}$, and $\tilde{\textbf{V}}=\tilde{\textbf{U}}^\transpose\matr{V}$.

It is known that the volume of simplex confined by the origin and the columns of $\tilde{\textbf{V}}$ can be computed by $\left|\det(\tilde{\textbf{V}})\right|$, where $\det(\tilde{\textbf{V}})$ is the determinant of $\tilde{\textbf{V}}$. As the distance from the origin to the \textit{sum-to-one hyperplane} is fixed, minimizing the volume of simplex that contains all \textit{utterance points} is equivalent to minimizing the volume of simplex defined by the origin and the columns of $\tilde{\textbf{V}}$, the MVS problem can thus be formulated as follows:
\begin{equation}
	\label{eq:eq5}
	\begin{split}
	    \tilde{\textbf{V}}^* = &\argmin_{\tilde{\textbf{V}}} \left|\det(\tilde{\textbf{V}})\right|\\
	    s.t. \quad \tilde{\textbf{V}}^{-1}\tilde{\textbf{P}} \succeq 0& \quad and \quad \textbf{1}_K^\transpose \tilde{\textbf{V}}^{-1}\tilde{\textbf{P}}=\textbf{1}_M^\transpose.
	\end{split}
\end{equation}

Let $\matr{Q}\equiv\tilde{\textbf{V}}^{-1}$, we have $\det(\matr{Q})=1/\det(\tilde{\textbf{V}})$. The optimization problem (\ref{eq:eq5}) can then be written as follows:
\begin{equation}
	\label{eq:eq6}
	\begin{split}
	    \textbf{Q}^* = &\argmax_{\textbf{Q}} \log\left|\det(\textbf{Q})\right|\\
	    s.t. \quad \matr{Q}\tilde{\textbf{P}} \succeq 0& \quad and \quad \textbf{1}_K^\transpose \matr{Q}\tilde{\textbf{P}}=\textbf{1}_M^\transpose.
	\end{split}
\end{equation}
Note that (\ref{eq:eq6}) is non-concave and may have many local maxima which makes it difficult to find the global optima. In \citep{li2008minimum}, the sequential quadratic programming (SQP) algorithm is proposed to find the sub-optimal solutions of (\ref{eq:eq6}) which will be described as follows.

We first simplify the constraint $\textbf{1}_K^\transpose \matr{Q}\tilde{\textbf{P}}=\textbf{1}_M^\transpose$. Note that every column of $\tilde{\matr{P}}$ can be represented as a linear combination of $K$ linearly independent vectors picked among the columns of $\tilde{\matr{P}}$: let the $K$ linearly independent vectors be expressed as the matrix $\matr{P}_K=[\tilde{\matr{p}}_{i_1} \; \ldots \; \tilde{\matr{p}}_{i_K}]$. Every column of $\tilde{\matr{P}}$ can be written as $\tilde{\matr{p}}=\matr{P}_K \beta$, where $\textbf{1}_K^\transpose\beta=1$. Consequently, we can re-write the constraint as the equality constraint $\textbf{1}_K^\transpose \matr{Q}=\textbf{1}_K^\transpose \matr{P}_K^{-1}\equiv \matr{c}$ where $\matr{c}=\textbf{1}^\transpose\matr{P}_K ^{-1}$ which is a constant row vector. Hence, the problem (\ref{eq:eq6}) becomes:
\begin{equation}
	\label{eq:eq7}
	\begin{split}
	    \textbf{Q}^* = &\argmax_{\textbf{Q}} \log\left|\det(\textbf{Q})\right|\\
	    s.t. \quad \matr{Q}\tilde{\textbf{P}} \succeq 0& \quad and \quad \textbf{1}_K^\transpose \matr{Q}=\matr{c}.
	\end{split}
\end{equation}
An SQP-based algorithm is presented in \citep{li2008minimum} to solve the problem in (\ref{eq:eq7}). The Vertex Component Analysis (VCA) algorithm \citep{nascimento2005vertex} is first applied for initialization. The output of VCA is a set of $K$ vectors from the columns of $\tilde{\matr{P}}$ the projected utterance points of the given corpus. To lighten the computation, we can discard the points inside the convex hull generated by the $K$ VCA vectors. To further speed up the algorithm, we can expand convex hull so that more points are inside and can thus be discarded. The details of the algorithm are shown in Algorithm 1.

\begin{algorithm}[H]
\KwInput{ Number of vertices $K$, Projected point set $\tilde{\textbf{P}}$, objective function $f(\matr{Q})\equiv \log|\det(\matr{Q})|$ }
\KwOutput{matrix $Q$}
\setstretch{1.35}
\begin{algorithmic}[1]
\State $\matr{M} := VCA(Y,K)$
\State $\matr{Q_o} := Expand(\matr{M})$
\State $\matr{Y} := Discard(\matr{Y});$ if $\matr{y}$ is inside the simplex spanned by $\matr{Q_o}$
\State Inequality constraint

$\matr{A} * \matr{Q} \geq \matr{b}$, $\matr{A} = \matr{Y}^T \otimes \matr{I}_K$, $\matr{b}=\matr{0}_{KM}$
\State Equality constraint

$\matr{Aeq}*\matr{Q}=\matr{beq}, \matr{Aeq}=\matr{I}_K \otimes \matr{I}^T_K, \matr{beq} =\matr{c}^T$\
\State $g(\matr{Q}) := -(\matr{Q}^{-1})^T, $ where $g(\matr{Q})$ is the gradient of $f$
\State $[H(\matr{Q})]_{i,j} := -[g(\matr{Q})_{:,j}*g(\matr{Q})_{i,:}]$, 

where $H(\matr{Q})$ is the Hessian matrix of $f$
\State $\matr{Q}:=SQP(f, \matr{Q}_o, \matr{A}, \matr{b}, \matr{Aeq}, \matr{beq}, g, H)$
\end{algorithmic}
\caption{Minimum Volume Simplex Analysis (MVSA)}
\end{algorithm}

\section{Model Properties} 
\label{sec:exp1}
In this section, we investigate the properties of CPM. We first use the geometric properties to derive \textit{semantic patterns}, which are structured compositions of semantic frames. Then we analyse the utterance-specific \textit{semantic structures} using visualization methods from the perspective of compositional analyses as well as word-word correlation analysis. 

All of the experiments are conducted on the same dataset as \citep{hakkani2016multi}, with 4,978 utterances from the training sets of the ATIS-2 and ATIS-3 corpora of the air travel domain \citep{price1990evaluation}. 
After embedding the utterances of the corpus into the low-dimensional affine subspace via PCA, we apply the algorithm proposed in \citep{li2008minimum} for generating the MVS-type convex polytope, implemented in MATLAB.
\subsection{Interpret Semantic Properties from Geometric Properties}
We demonstrate the process of interpreting the \textit{geometric properties} of the generated convex polytope from the \textit{semantic perspective} with two convex polytopes with different dimensions: one two-dimensional (2-D) polytope to visualize the result and one higher dimensional (75-D) polytope to explore our framework's ability. The patterns extracted from geometric properties of the polytopes show great potential in exhibiting the semantic-frame-related information(i.e., intents and related slots).
\subsubsection{Two-dimensional polytope analyses}
\begin{figure}
    \centering
    \subfloat[The 2-D MVS-type polytope. Vertices(circled) are labelled as V1-V3. The scattered dots denote projected utterance points.]{
    \includegraphics[width=0.45\linewidth,valign=B]{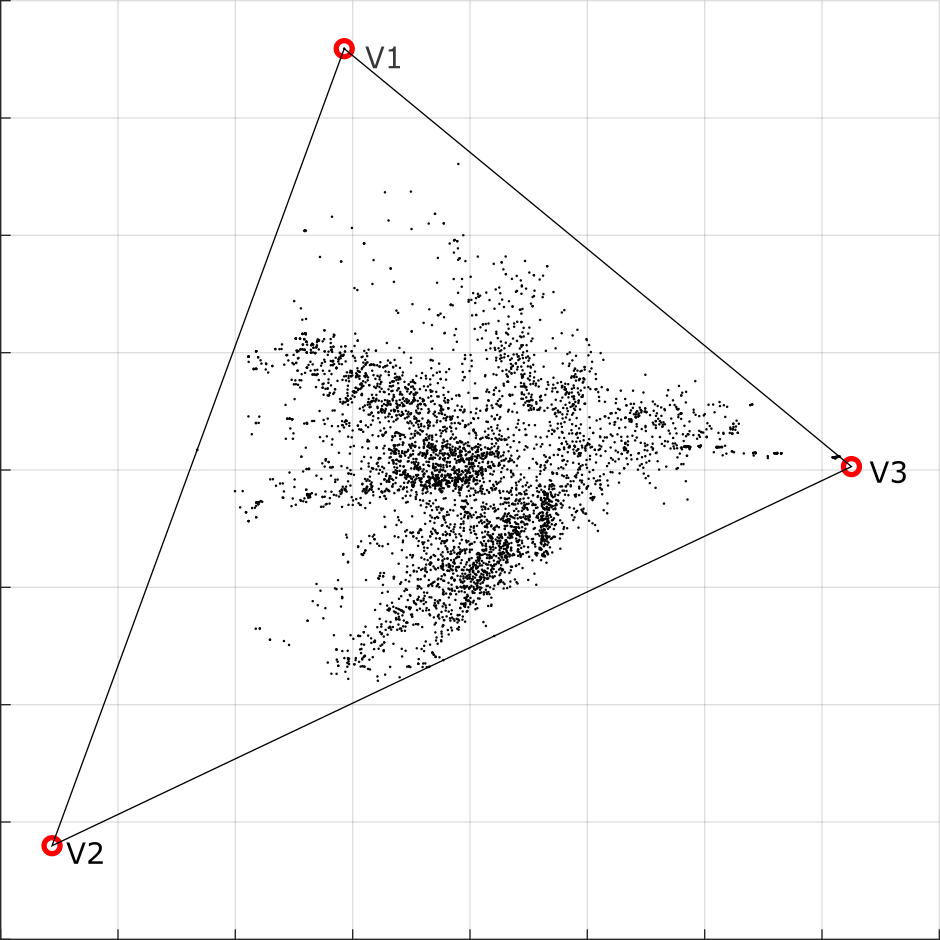}}
   \quad
   \subfloat[Nearest utterances and top terms corresponding to the vertices in the 2-D MVS-type polytope.]{\includegraphics[width=0.45\linewidth,valign=B]{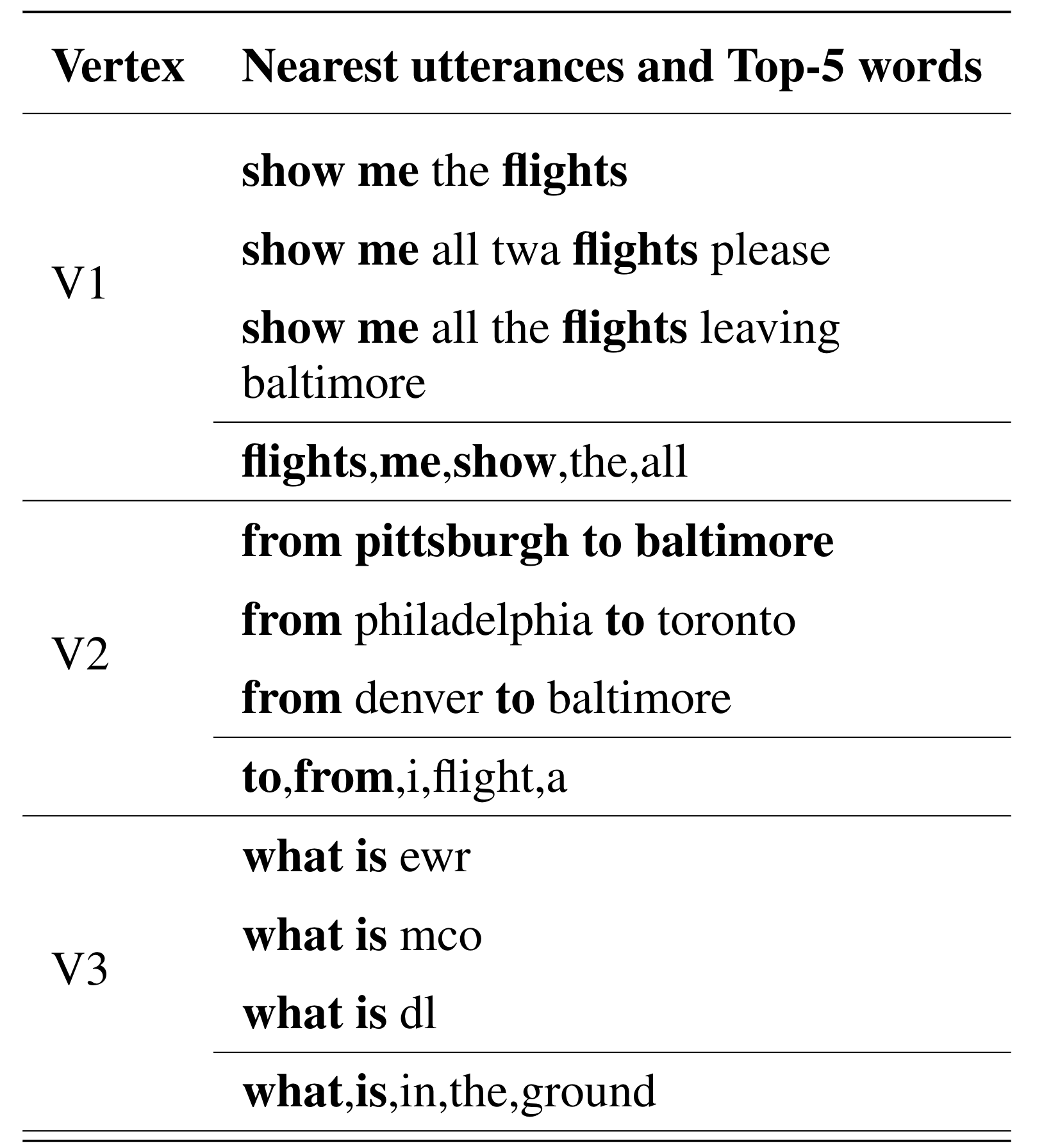}} 
   \\
     
    \caption{Results corresponding to the 2-D MVS-type polytope.}
    \label{fig:2d-mvs}
\end{figure}

We start our investigation of the geometric properties from a two-dimensional (2-D) experiment (i.e., finding minimal volume simplex (MVS) with three vertices on a 2-D subspace) for better visualization. 
After transforming the corpus into a sum-normalized term-utterance matrix, the \textbf{utterances} are represented as \textbf{data points} in the original coordinate system where the axes correspond to the terms in the corpus. 
Then we project all the utterance points onto a 2-D affine subspace via PCA and generate the MVS as shown in Fig. \ref{fig:2d-mvs}a. 
The scattered dots in Fig. \ref{fig:2d-mvs}a denote the projected utterance points. The three circled points marked as $V1$, $V2$, and $V3$ are the \textbf{vertices} of the polytope, which are the extracted \textbf{semantic patterns} covered by the minimal volume simplex. 
The scope of the dialog corpus is roughly defined by the \textbf{simplex}.

Next, we focus on interpreting the semantic patterns. 
In Section \ref{sec:cp}, we have discussed that the semantic patterns in the corpus can be characterized by the vertices $\matr{V}$ of the polytope, and utterances are further represented by the convex combination $\alpha$ of the vertices. 
Here we extend the previous work in \citep{zhou2021automatic} and use two geometric properties of $\matr{V}$, namely, \textit{point coordinates} and \textit{Euclidean distances}, to interpret the semantic patterns in the dataset extracted by the model.
First, in the original coordinate system, each axis corresponds to a term. For the utterance points, its value on each axis represents the contribution of the corresponding term to that utterance. Similarly, for a vertex $v$ representing a specific semantic pattern, its coordinate values capture the contributions of the related terms to that pattern. Thus, the top-$k$ terms of a vertex essentially represent the semantics captured by the vertex.
Second, the utterance points near each one another (from the Euclidean distance perspective) usually exhibit similar semantic patterns. Hence, we also utilize the $n$-nearest utterances of the vertices for interpretation.
The 3-nearest neighbouring utterances and top-5 terms of the vertices are listed in Fig. \ref{fig:2d-mvs}b. 

The three vertices are related to clear semantic patterns indicated by their top terms and nearest neighbouring utterances (marked in boldface in Fig. \ref{fig:2d-mvs}b ). Namely, the pattern of $V1$ is: `show me ... flights'; $V2$ abstracts `from \textit{city-name} to \textit{city-name}'; and $V3$ represents `what is \textit{code}'. Again the tokens that belong to the same semantic type are delexicalized with generalized labels including \textit{city-name}, \textit{code}, etc.

The above observations indicate that the vertices of the MVS-type polytope can capture distinctive and representative semantic patterns, which are interpretable through analysing the closest neighbouring utterances and top terms. Furthermore, these patterns associate closely with the structure of the semantic frames. The pattern captured by $V1$ indicates the intent \textit{FLIGHTS}, the pattern corresponding to $V2$ contains the slot information \textsc{origin} and \textsc{destination}, and  the pattern corresponding to $V3$ reveals both the intent \textit{ABBREVIATION} and slot \textsc{airline-code}. 

\subsubsection{Higher-dimensional polytope analyses}
\label{sec:hda}
The actual possible patterns in the corpus should be far more than three. Higher dimensional polytope can capture more semantic patterns. But too many vertices can result in repeating patterns. Therefore, we attempt to analyse the CPM at higher dimensionality. After observing the results of many different dimensions, a 75-dimensional polytope is selected as the best dimensionality for this corpus to extract as many distinctive patterns as possible.

Analysis in the 75-dimensional space focuses also on the vertices of the convex polytope. The utterances closest to the vertices are analysed, together with their most frequent terms.
More specifically, we automatically interpret the extracted patterns by searching for the most popular combinations of terms in the top-10 term list among the 50-nearest neighbouring utterances of the corresponding vertices. 
We list part of the patterns exhibited by the vertices of the 75-dimensional polytope in Table \ref{tab:mvs-75d}. We also located the patterns in the utterances and compared them with the semantic-frame labels. The related intent and slot tags are also presented in the table. 
In addition to the semantic patterns that have already been found in the two-dimensional analyses above (namely, $V5$, $V10$ and $V24$ in Table \ref{tab:mvs-75d}), we also found many new semantic patterns that are related to the users' intent, as well as slot information. For example, $V17$: `at  \textit{airport-name} airport' indicates the slot \textsc{airport-name}, and $V51$: `type of' is closely related to the intent querying \textit{TRANSPORT TYPE}. 
\begin{table}[h]

\caption{Examples of extracted semantic patterns and related slot tags found in the 75-dimensional MVS-type convex polytope. The tags written in capitals are intent tags, and slot tags are formatted in small capitals.}
\centering
\begin{tabular}{ccc}
\toprule
\textbf{Vertex}	& \textbf{Extracted Semantic Patterns}	& \textbf{Related Intent/Slot Tags}\\
\midrule
V1 & one way ticket/fares	& \textsc{round-trip, cost-relative}\\
V5 & from \textit{city-name} to \textit{city-name} & \textsc{fromloc.city-name, toloc.city-name}\\
V6 & fare code \textit{code} & \textsc{fare-basis-code}\\
V10 & show me \textit{airport-name}/\textit{airline-name} & \textsc{airport-name, airline-name}\\
V17 & at \textit{airport-name} (airport) & \textsc{airport-name}\\
V24 & what is ... \textit{code} &  \textit{ABBREVIATION}, \textsc{airline-code}\\
V36 & ground transportation at/in  &  \textit{GROUND SERVICE} \\
V42 & round trip fares  & \textsc{round-trip, cost-relative}\\
V48 & between \textit{city-name} and \textit{city-name}  & \textsc{fromloc.city-name, toloc-city-name}\\
V51 & type/kind of  & \textit{TRANSPORT TYPE/AIRCRAFT} \\
V69 &first class & \textsc{class-type}\\
... & ... & ... \\
\bottomrule
\end{tabular}
\label{tab:mvs-75d}
\end{table}

Aside from observing the patterns directly, we also conduct quantitative analyses on the relationship between the vertices and semantic-frame-related information (i.e., intents and slots).
The extracted patterns cover \textbf{82.5$\%$ (66/80)} of all the slot categories found in the labelled data. Also, the patterns or combination of patterns can clearly reflect information about the user intents. 
 Note that semantic frames with different user intents usually contain different slot information. Thus the utterance points with different intents should associate with different sets of vertices. For example, for user utterance point with intent \textit{GROUND SERVICE}, it is usually composed of vertices including  $V35$: `ground transportation', $V17$: `at \textit{airport-name} (airport)', and $V33$: `give me', e.g., `\textit{show me the ground transportation in Denver}'. While for utterance points with intent \textit{CAPACITY}, the most frequently appearing vertices includes $V2$: `how many' or $V52$: `capacity of', and $V24$: `what is ... \textit{code}', e.g., `what is the capacity of an f28' where `f28' is a \textit{code}.

\begin{figure}
    \includegraphics[width=\textwidth,height=\textheight,keepaspectratio]{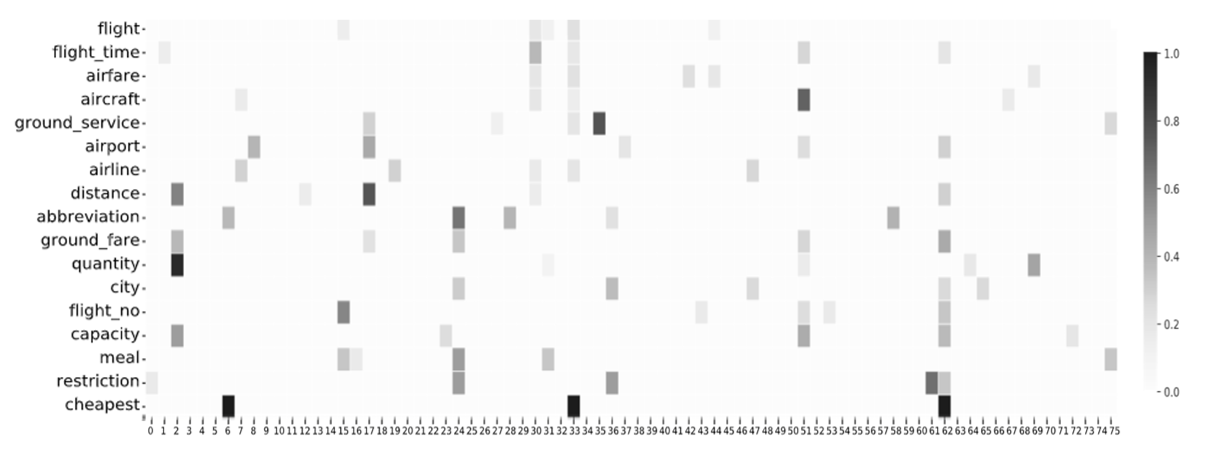}
    \caption{Heatmap showing the frequency of occurrence of vertices in each intent. The horizontal axis shows the vertices and the vertical axis shows the intents. The shade of the block indicates the frequency of occurrence of the vertex in the utterance point set with corresponding intent. For each intent, only the five most frequently appearing vertices are plotted.}
    \label{fig:intent}
\end{figure}
Here we plot a heatmap (shown in Fig. \ref{fig:intent}) to check the frequency of occurrence of vertices in the utterance sets with the same intents. We first categorize the user utterances into different sets according to their intents. Then we calculate and plot the frequency of the five most frequently appearing vertices in the utterance set for each intent. 
Specifically, for intent $i$, we have a set of utterances $U_i$ = ${u_{i_1},u_{i_2}, u_{i_3}, ... , u_{i_n}}$ with this intent. Based on our previous discussion, each utterance point $u_{i_j}$ can be represented as a convex coefficient vector  $\matr{a}_{i_j}$, which measures the contribution of each vertex to the point. To reduce noise, for each utterance $u_{i_j}$ we only consider the three most important vertices as `appearing', i.e., the vertices with corresponding convex coefficients rank top three in $\matr{a}_{i_j}$. Then we can get the frequency of the vertices by calculating the number of occurrences in the whole utterance set $U_i$, which is indicated by the shade of the blocks in Fig. \ref{fig:intent}.

The result in Fig. \ref{fig:intent} shows that no two intents have the same vertices combination, which is consistent with our observation that they are associated with different slot-value pairs.  Also, most of the intents have high frequency (higher than $50\%$) vertices in the utterance set, which means that most of the utterances with this intent contain the patterns exhibited by these vertices.  Therefore, we can conclude that the intent information of the user utterance is contained in the extracted features. 
Our following experiments are all based on the 75-dimensional convex polytope result for consistency.

\subsection{Utterance-specific structural analyses: Case study by visualization}
\label{sec:coeff}

After interpreting and analysing the vertices extracted by CPM, we now conduct utterance-specific analyses using the vertices and convex coefficients. 

\subsubsection{Compositional analyses and visualization}
\label{sec:exp2.1}
As mentioned in Section \ref{sec:cp}, the utterance points can be decomposed as the convex combination $\matr{a}$ of vertices $\matr{V}$. 
Our previous experiments demonstrate that the vertices can be interpreted as semantic patterns that contain intent and slot information.  
By analysing the composition of patterns in each utterance, we can utilize the patterns to understand the queries in terms of their intents and slots. 
Following this argument, we look at the algebraic compositional representation of the utterance points: convex combination $\matr{a}$. We now show that utterance points with similar semantic frames share a similar combination of vertices.

\begin{figure}[thb]
    \centering
    \includegraphics[width=1.03\textwidth,height=\textheight,keepaspectratio]{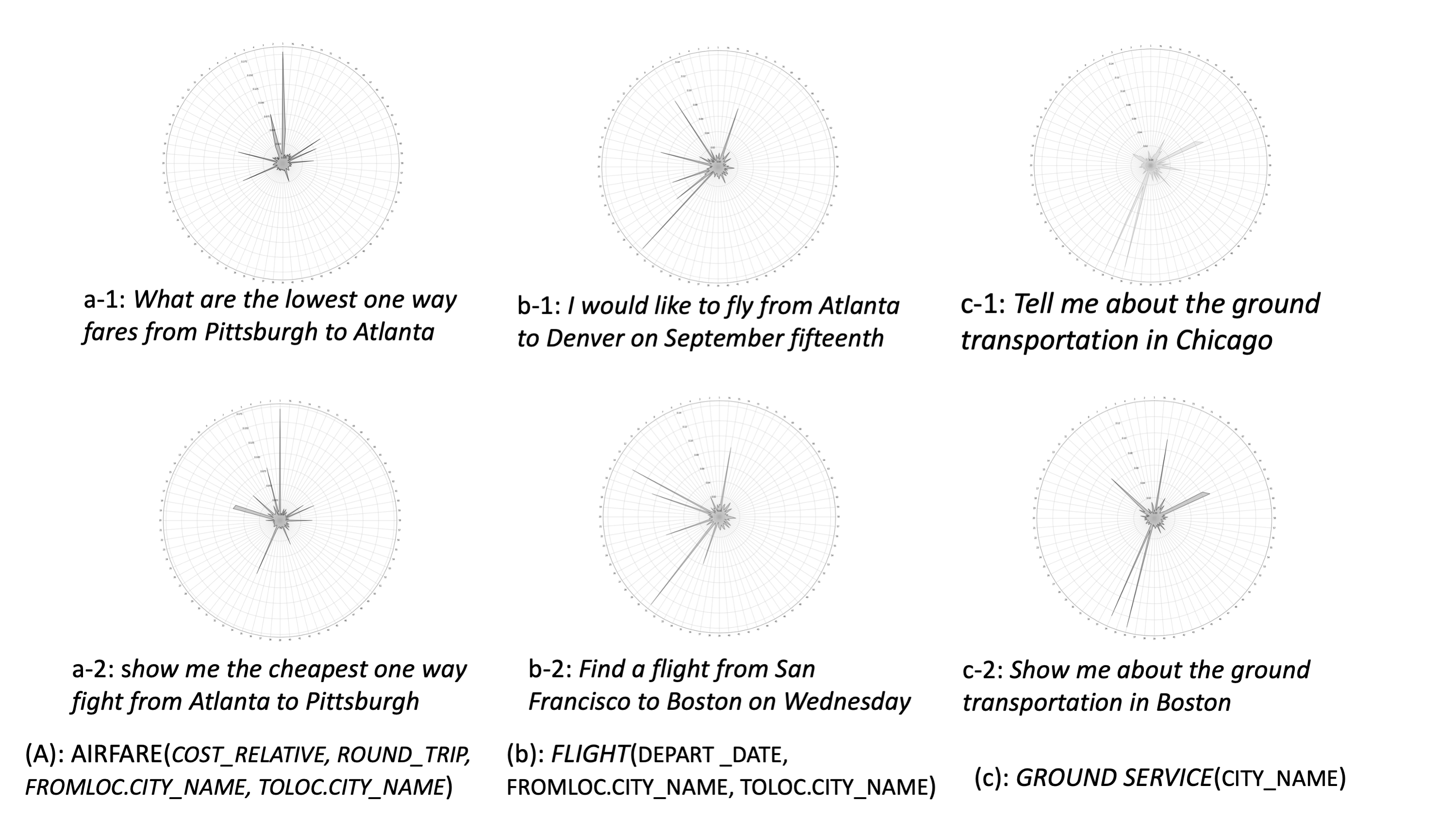}
    \caption{Radar graphs of three groups (a, b, and c) of utterances with the same intent and slot tags, which are labelled in the group titles. The axes in the radar graph are the vertices of the polytope. The coordinate values
are the coefficients of the vertices of each utterance. }
    \label{fig:radars}
\end{figure}

\textbf{Settings.} 
To visualize the portion of different vertices in the utterance, we plot the radar graphs of the convex coefficients of the utterance points (shown in Fig. \ref{fig:radars}) . In the radar graphs, each axis represents one vertex, and the 76 vertices are arranged anti-clockwise. The coordinate values are the coefficients $a$ of the vertices of each utterance. 
We plot the radar graphs of three groups of utterances. Each utterance group contains two utterances with the same intent and slot tags (note that their slot values are not necessarily  the same), which are labelled in the title of the group. 
\textbf{Results and analyses.}
The results in Fig. \ref{fig:radars} demonstrate that utterances with the same intent and slot tags show a high degree of consistency of their radar graph, although they use different terms and phrases to convey their goals. 
Although the graphs of each group are not exactly the same, there is a large overlap of their most important vertices (i.e., vertices that have the largest coefficients). 
For example, the utterances in (c) are seeking for \textit{GROUND SERVICE} information at certain cities. The three most important vertices that two utterances have in common are: $V36$: `ground transportation', $V34$: `give/tell me', and $V63$: `arriving/in \textit{city-name}'. 
The utterances in the other two groups also show similar shapes of their radar graphs.
Therefore, the convex coefficients of utterance points indicate their relationships with the vertices and thus contain the knowledge of semantic frames of the utterances.


\subsubsection{Word-word correlation analyses and visualization}
\label{sec:exp2.2}
 
In the above, we have validated that different compositions of vertices indicate different semantic frames in utterances. We now turn to exploring the underlying semantic structures of the utterances. 
In Section \ref{sec:exp1}, the vertices are demonstrated to be closely related to semantic patterns, and the patterns can be discovered by examining the \textit{top terms} of the vertices. Hence, different terms should associate with different semantic patterns conversely. 
Therefore, we utilize the extracted vertices to generate the word representations, which is expected to contain sufficient semantic-frame-related information. 
Then we use this CPM-based word representation to calculate the word-word correlation for each utterance and visualize the underlying semantic structures in the utterances. 

\textbf{Settings.}
Each column of vertex matrix $\matr{V}^{\in N \times K}$ represents the coordinates of a vertex.  From the perspective of \textbf{row vectors}, each entry in the $i^{th}$ \textbf{row} of vertex matrix $\matr{V}$ indicates the correlation between the vertices and the word $t_i$.
Therefore, each word can be represented by a vector $\matr{V}^{T}_{i}{\in K}$, which is a  column vector and encodes the semantic-frame-related information. As the entries in this CPM-based word vector are closely related to semantic patterns, words that belong to similar semantic patterns have similar vectors.
Then, we apply the standard cosine similarity of two-word vectors to obtain word-word correlation matrix $\matr{M}$ for each utterance:
\begin{equation}
    \matr{M}_{[i,j]} = \frac{\matr{V}^{T}_{i} \matr{V}^{T}_{j}}{\norm{\matr{V}^{T}_{i}} \norm{\matr{V}^{T}_{j}}}
    \label{eq:coss}
\end{equation}
Its entry $\matr{M}_{[i,j]}$ denotes the similarity between $i$-th and $j$-th words in the utterance. 

\begin{figure}
\includegraphics[scale=.5]{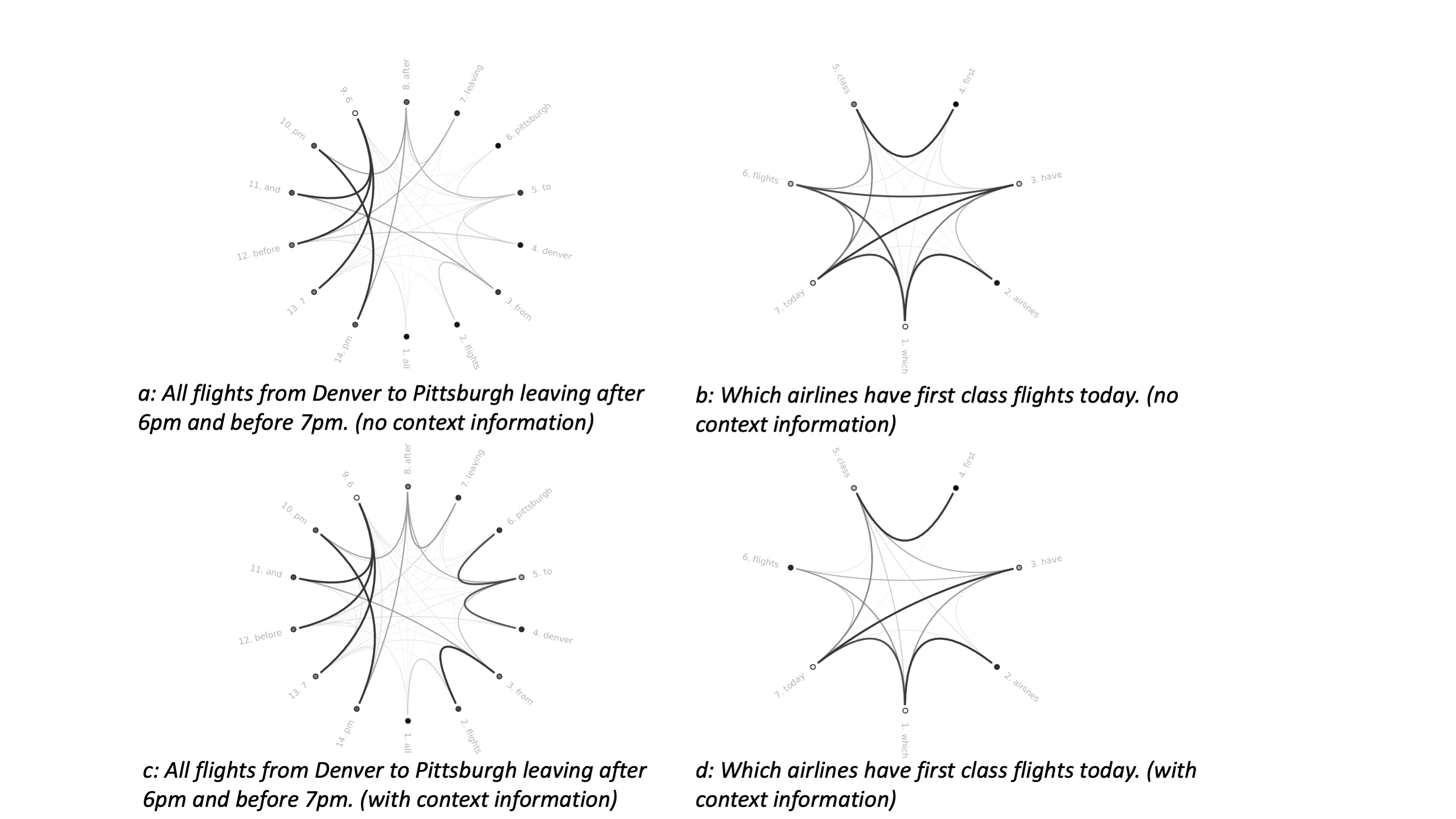}
   \\
    \caption{Schema-balls of word-word correlation in the utterances (a, b: no contextual information; c, d: with contextual information). Nodes are words in the utterance and are arranged anti-clockwise. The thickness of the line between two nodes indicates the correlation strength between two words from the perspective of semantic patterns.}
    \label{fig:uni-sechma}
\end{figure}

Current word-word correlation can be derived from the terms' affinity with vertices closely related to semantic patterns. However, the generation of vertices is based on a bag-of-word model, which discards the ordering information in the corpus. To include the sequential information, we propose a contextual word-word correlation that takes the surrounding words (i.e., contexts)  into consideration. We set a window size $w$: when calculating one word $t_m$'s correlation with the other word $t_n$, the correlation between words within the window and the target word $t_n$ are also considered.  Formally, we set window size as 3, i.e., we put the word immediately preceding and succeeding the target word $t_m$ into the window. We calculate the new correlation matrix $\matr{M}_c$ as follows:
\begin{equation}
    \matr{M}_{c_{[i,j]}} = \matr{M}_{[i,j]} + \matr{M}_{[i-1,i]} \matr{M}_{[i-1,j]} + \matr{M}_{[i+1,i]}\matr{M}_{[i+1,j]}
    \label{eq:context}
\end{equation}
The contextual word-word correlation between $i$ and $j$ is a weighted sum of the correlation between $\{i-1, i, i+1\}$ and $j$ where the weights for the contextual words is the correlation between them and the main word $i$ (i.e., $\matr{M}_{[i+1,i]}$ and $\matr{M}_{[i-1,i]}$). 
For each utterance, we use the schema-ball to visualize the word-word correlation in the utterance. Each word in an utterance is represented as a node in the schema-ball, and the words are arranged anti-clockwise. We also added serial numbers before the words to indicate their position in the utterances for better illustration. The thickness of the line between two nodes indicates the strength of the relationship between the two words. We use two utterances as examples and their schema-balls generated by $M$ and $M_c$ are shown in Fig. \ref{fig:uni-sechma}.

\textbf{Results and analyses.}
By connecting the words with high correlation indicated by bold lines in the schema-ball, we can reconstruct the underlying semantic structure of the utterances. The semantic structure shown by the schema-ball in Fig.\ref{fig:uni-sechma}a reveals the semantic pattern `after 6 pm and before 7 pm' in the utterance, which indicates the slot \textsc{time} information. However, it ignores other structures including `all flights from' and `from Denver to Pittsburgh', which are captured by the contextual word-word correlation graph in Fig. \ref{fig:uni-sechma}c. 
Fig. \ref{fig:uni-sechma}a shows (i) `which airlines', (ii) `first class' and (iii) the relation between `today'  and other words. Aside from above information, the contextual word-word correlation in Fig. \ref{fig:uni-sechma}d also captures the relation between `flights' and other words including `first class flights', `which flights', `have flights', etc. 
The results above show that the CPM-based word representation can indicate their semantic role in the utterances (i.e., words that form a semantic pattern tend to have a higher correlation). Also, adding contextual information can find more reasonable semantic structures for understanding the utterance. 
\section{CPM-Guided Neural Model for NLU}
\label{sec:nn}
In this section, we will demonstrate that CPM-extracted features can assist NLU tasks at various levels , and bring improved interpretability and transparency to NLU models.  Our investigation is based on the slot filling task on ATIS dataset as showcases, where features related to the semantic frame, as mentioned in the experiments above, are used as additional input to neural network models. We expect the features will help the models achieve higher per- formance in language understanding tasks, while being interpretable as they are derived from a transparent CPM generation process.

\subsection{Background}
\begin{figure}[h]
    \centering
    \includegraphics[width=\linewidth]{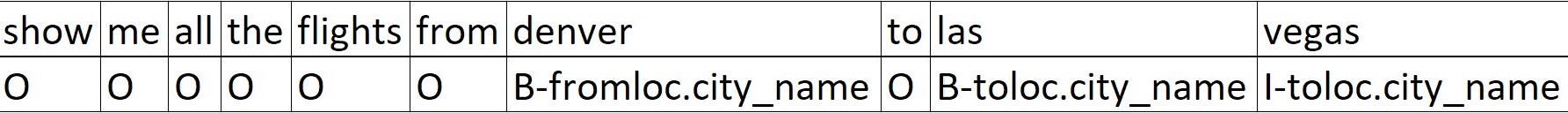}
    \caption{An example utterance and its corresponding slot label sequence in ATIS dataset}
    \label{fig:frame}
\end{figure}

\textbf{Task: Slot filling} The core of slot filling is sequence labelling:  input a word sequence $\mathbf{x} = (x_1, \ldots, x_n)$, the model aims to map it to a slot label sequence $\mathbf{y} = (y_1, \ldots, y_n)$. 
The slots are tagged in the `IOB' style as in \citep{hakkani2016multi}.
Fig. \ref{fig:frame} shows an example of a slot label sequence for an utterance in the ATIS dataset. 

\noindent\textbf{Base model: RNN encoder-decoder} We use the model structure in \citep{liu2016attention} to complete the slot filling task. The encoder is a single-layer bi-directional RNN. The decoder is a single-layer uni-directional forward RNN to predict slot labels sequentially. LSTM cells, described in \citep{10.1162/neco.1997.9.8.1735}, are used as the fundamental unit for RNN layers in both the encoder and decoder. The initial hidden state of the decoder is the last state of the encoder RNN. At each time step $i$, the decoder takes the label predicted at the previous step: $y_{i-1}$, along with the related encoded feature vector at the same timestep: $h_i$, and a context vector $c_i$ to predict the distribution of $y_{i}$. The context vector $c$ is calculated by a weighted addition of feature vectors emitted by \textit{all} the encoder states, i.e.:
\begin{equation}
    c_i = \sum_{j=1}^n a_{i,j} h_j
\end{equation}
where $a_{i,j}$ is the weight (i.e., attention) applied to feature vector at timestep $j$ when decoding timestep $i$. This attention mechanism takes the relationships among the words into consideration when decoding the slot labels.
Conventionally, the weight is learnt through a fully-connected layer $g$ (Eq. \ref{eq:fclayer}), then normalized to satisfy $\sum_{j=1}^n a_{i,j} = 1$ (Eq. \ref{eq:eq14}).
\begin{equation}
\begin{aligned}
    e_{i,k} &= g(s_{i-1},h_{k})
    \label{eq:fclayer}
\end{aligned}
\end{equation}
\begin{equation}
\begin{aligned}
    a_{i,j} &= \frac{\exp(e_{i,j})}{\sum_{k=1}^n \exp(e_{i,k})} 
    \label{eq:eq14}
\end{aligned}
\end{equation}
This base model setting is used as baseline in the experiment.
\subsection{Proposed method: CPM-guided slot filling}
We have illustrated in Section \ref{sec:coeff} that the word-word correlation matrix $\matr{M}_c$ generated by CPM reveals the semantic structures of the utterances. Note that $a_{i,j}$ is also expected to denote the correlation between words.
Therefore, by adapting the matrix $\matr{M}_c$ as the attention to the RNN decoder, we can incorporate the semantic-frame knowledge extracted from the corpus by CPM into the neural language model for slot filling. In the base model, the fully-connected layer learns appropriate attention of one position attributed to other positions in an utterance, to reflect that importance of semantic information at different positions is not equal when understanding and predicting slot information at a particular position. However, this relationship has already been computed by CPM: the experiment in section \ref{sec:exp2.2} has shown that the word-word correlation in entries of matrix $\matr{M}_c$ is closely related to their semantic role in the utterances, and when decoding slot label, using it as attention guides the decoder into applying more weight on information at positions with closer affinity in terms of semantic structures of the utterance.

We use the same structure of the base model, but instead of adopting a fully-connected layer, we first derive the word-word correlation matrix $\matr{M}$ and $\matr{M}_c$ as shown in Experiment \ref{sec:exp2.2} before training the RNN encoder-decoder. All the negative values are treated as an indication of no correlation and are zeroed out. The row vectors are then normalized as sum-to-one vectors and applied as attention vectors for the decoder. Other structures of base model are kept the same as baseline.
In summary, two methods are compared in this experiment, differentiated by their attention $a_{i,j}$ generation mechanism:
\begin{enumerate}
    \item \textbf{Baseline.} The attention is learnt by the fully-connected layer as described in Eq. \ref{eq:fclayer}.
    \item \textbf{CPM-guided.} The attention is from entries in the word-word correlation matrix with contextual information $\matr{M}_c$ as shown in Eq. \ref{eq:context}.
\end{enumerate}
Note that in baseline model, the attention for words in utterances is generated during the training stage, while in our proposed CPM-guided model, it is computed before training the neural model, and not updated further once it is generated.
This experiment uses the same ATIS dataset as configured in previous section \ref{sec:exp1}.
To investigate the performance of our proposed mechanisms on varied sizes of annotated data, we configure three sets of experiments, in which the labelled training set size is 2.5\%, 5\%, and 90\% of the whole training set, respectively. We use a validation set containing 500 utterances for all the experiments.  Because generating convex polytope on a corpus does not require any labels at all, the CPM generation process uses the whole training set regardless of the labelled data size for the latter neural base model.

The CPM-generated attention vectors are not updated throughout the experiment since they are derived during the generation of CPM on the corpus beforehand, not the training of neural slot filling model. In contrast, we use the aforementioned fully-connected layer to learn attention on the fly, and the parameters of which are allowed to be updated throughout the training phase.

The configuration for base model is as follows: Words in utterances are represented with a trainable embedding layer, randomly initialized and the dimension of the word embedding is 150. 
Since the slot labels are imbalanced, we use focal loss explained in \citep{DBLP:journals/corr/abs-1708-02002} as the loss function. 
Models are tuned on validation sets among possible hyperparameter combinations:  (i) the output dimension of encoder LSTM cell is in \{50,100,150,200\} and the dimension of decoder LSTM cell output is kept at 2 times of the setting for encoder cells, (ii) the dropout is chosen from \{0.2,0.3,0.4,0.5\}, (iii) the gamma parameter for focal loss is in \{0,1,2,3\}, and (iv) the batch size is selected from \{1,2,4,8,16,32,64\}. We use Adam as the optimizer. \cite{DBLP:journals/corr/KingmaB14} For evaluation, F1 score is calculated based on the correctly identified slot chunks in slot filling prediction result on the test set.

\subsection{Results and analyses}
\begin{table}[h]
\centering
\begin{tabular}{c  c  c  c  c}

\hline
    
    \multirow{2}{*}{Model} & \multicolumn{4}{c}{Data size} \\
    \cline{2-5}
    & 2.5\% & 5\% & 10\% & 90\% \\
    \hline
        Baseline & 52.39 & 72.22 & 79.16 & 94.24 \\   
        \hline
        CPM-guided & \textbf{66.07} & \textbf{75.37} & \textbf{83.48} & \textbf{94.60} \\
        \hline
        \hline
        F1 Increase & \textbf{26.11\%} &4.36\% &5.46\%&0.38\%
        \\

    \hline
\end{tabular}
\caption{F1 score of predicted slot labels on different training data size (in percentage of 4978 training utterances) and different models. Our proposed mechanism achieve better slot prediction accuracy (results marked in bold) on all sizes of training dataset, up to 26\% of the performance on baseline model.}
\label{tab:att}
\end{table}
Table \ref{tab:att} shows the slot filling F1 scores of our experiments on varied sizes of training set.
Compared with the baseline,
models guided by CPM-generated attentions  exhibit significant improvements on 2.5\%, 5\% and 10\% data size. 
The result demonstrates strength of CPM-guided model from two perspectives: First, CPM can process labelled and unlabelled utterances and extract semantic knowledge, while the neural networks can only learn from annotated data. Thus by incorporating the external semantic knowledge, CPM-guided model can outperform the baseline model. This is a favourable feature as an addition to the use of neural networks, which may overfit when the labelled training data size is limited, but more unlabelled data within the same domain could be easily obtained.
Second, CPM attention is interpretable by the semantic patterns we discussed in previous sections, and the reproducible generation process is explicitly defined stepwise, while learning attention through fully-connected layer is not easily interpretable, since it is difficult to relate the kernel, bias and activation function in the layer to semantic features. 
Moreover, the prohibitive amount of parameters in kernel and bias matrix and non-linear transformation in activation function is not directly perceivable by human, hence opaque from human comprehension.
Additionally, with pre-computed attention, CPM-guided model saves computational resources.

In summary, CPM features at the word-level (word-vertex matrix in Section \ref{sec:exp2.2}) and utterance-level (convex combination coefficients discussed in \ref{sec:exp2.1}) are closely related to semantic frames, as visualized, and further validated by using them in slot filling. When used as guidance in neural network models, these features have a positive impact on the model performance, while not losing transparency: all of the added features into neural networks are interpretable through understanding the CPM construction process.
\section{Conclusions and Future Work}
This paper proposes a novel, unsupervised, data-driven framework based on the Convex Polytopic Model (CPM) to discover the semantic structures in raw conversational data, as well as provide an data efficient approach for NLU. This framework extracts key semantic patterns from raw task-oriented dialog corpus, discovers the underlying semantic structure of the utterances, and provide features related to semantic structures that can be ingested into neural language models.

Our experiments show that the automatically extracted patterns are strongly related to the intent and slot-value pairs in the semantic frames. Also, the compositional analyses and the underlying structure discovery agrees with the theory that utterances could be convexly represented by a combination of the MVS vertices. We have demonstrated the generation and application of the geometric properties of the extracted semantic features, the process of which is concisely and concretely defined, and the results are easily understood. 
Stemming from the extracted features, we approach slot filling task on ATIS dataset with CPM features as additional input to NLU models. When the amount of labelled data is limited, our approach outperforms the baseline model by a large margin (up to 26\% F1 score increase). The result shows that CPM-extracted features boost the performance of the state-of-the-art approaches on the slot filling task with its interpretable, transparently generated features.

Future work will focus on extending the application of the CPM-based framework and utilizing it to more tasks in task-oriented dialog systems.

\section{Acknowledgements}
This work is partially supported by the General Research Fund from the Research Grants Council of Hong Kong SAR Government (Project No. 14245316). 

\bibliographystyle{nlelike}
\bibliography{References}
\label{lastpage}

\end{document}